# Data Valuation for Medical Imaging Using Shapley Value: Application on A Large-scale Chest X-ray Dataset

Siyi Tang[1], Amirata Ghorbani[1], Rikiya Yamashita[2], Sameer Rehman[3], Jared A. Dunnmon[4], James Zou[1,2,4], Daniel L. Rubin[*2,3]

[1]Department of Electrical Engineering, Stanford University, Stanford, CA, USA
[2]Department of Biomedical Data Science, Stanford University, Stanford, CA, USA
[3]Department of Radiology, Stanford University, Stanford, CA, USA
[4]Department of Computer Science, Stanford University, Stanford, CA, USA

[*]**Address correspondence to:**

    Daniel L. Rubin
    Professor of Biomedical Data Science and Radiology
    Stanford University
    Email: dlrubin@stanford.edu

**Abstract**

  The reliability of machine learning models can be compromised when trained on low quality data. Many large-scale medical imaging datasets contain low quality labels extracted from sources such as medical reports. Moreover, images within a dataset may have heterogeneous quality due to artifacts and biases arising from equipment or measurement errors. Therefore, algorithms that can automatically identify low quality data are highly desired. In this study, we used data Shapley, a data valuation metric, to quantify the value of training data to the performance of a pneumonia detection algorithm in a large chest X-ray dataset. We characterized the effectiveness of data Shapley in identifying low quality versus valuable data for pneumonia detection. We found that removing training data with high Shapley values decreased the pneumonia detection performance, whereas removing data with low Shapley values improved the model performance. Furthermore, there were more mislabeled examples in low Shapley value data and more true pneumonia cases in high Shapley value data. Our results suggest that low Shapley value indicates mislabeled or poor quality images, whereas high Shapley value indicates data that are valuable for pneumonia detection. Our method can serve as a framework for using data Shapley to denoise large-scale medical imaging datasets.

# Introduction

Modern machine learning methods such as deep learning have achieved impressive empirical performance in diverse medical image analysis tasks, including assessment of cardiac function from ultrasound[1], segmentation of neuronal structures from electron microscopy[2], skin lesion classification from dermatoscopy[3], intracranial hemorrhage detection from computed tomography[4,5], and automated chest X-ray interpretations[6,7]. These successes were generally made possible by the availability of large-scale hand-labeled training datasets. However, hand-labeling of medical images is particularly resource- and time-intensive because it relies on domain expertise[8]. An emerging alternative is to use crowd-sourcing or automated algorithms to label massive datasets, such as ChestX-ray14[9], DeepLesion[10] and MIMIC-CXR[11]. However, this resulted in less accurate labels compared to hand-labeled datasets[12,13]. Furthermore, in addition to inaccurate labels, images within a dataset might have noise and different types of artifacts and biases due to equipment or measurement errors[14]. Therefore, data quality in medical imaging datasets is a recognized challenge[15]. Low quality data can reduce the performance of machine learning models and may explain why models failed to generalize to new data from external sites[16]. Hence, algorithms that can automatically identify low quality data could help mitigate this problem. In this work, we propose using data Shapley[17,18], a state-of-the-art data valuation metric, to quantify the value of individual datum in a large chest X-ray dataset[9].

Several methods have been developed to handle inaccurate labels in medical imaging datasets[19], including adding a noise layer in the network architecture[20], data re-weighting[21,22], semi-supervised learning[23] and weakly supervised learning[24,25]. For example, previous works handled label noise in ChestX-ray14 dataset through model confidence[26] or graph-based semi-supervised learning[27]. Furthermore, denoising medical images has been studied extensively, ranging from classical techniques such as filtering[28,29] and wavelet transformation[30,31] to more recent deep learning methods[32–34]. Most of these prior studies focus on approaches to handling inaccurate labels or suboptimal images in model development or training stages. In contrast, our approach can directly identify low quality data, and thus allows us to clean up the datasets and improve the training of machine learning models.

Given a training set, a supervised learning algorithm and a predictor performance score, data Shapley[17] is a metric that quantifies the value of each training datum to the predictor performance. Experiments on small to moderate-scale biomedical, imaging and synthetic data have demonstrated that low Shapley value captures outliers and corruptions, while high Shapley value informs the type of new data that should be acquired to most efficiently improve the predictor performance[17]. Furthermore, data Shapley has been shown to substantially outperform the leave-one-out (LOO) score[35], a commonly used statistical measure of data importance. Moreover, data Shapley has several advantages as a data valuation framework[17]: (a) it is directly interpretable because it assigns a single value score to each data point and (b) it satisfies natural properties of equitable data valuation. To the best of our knowledge, our work is the first to apply data Shapley to biomedical imaging data. We aim to assess the effectiveness of data Shapley in capturing low quality data as well as informing valuable data in the context of pneumonia detection from chest X-ray images.

Our main contributions are as follows. First, we propose a framework (see Figure 1) to quantify the value of training data in a large chest X-ray dataset in the context of pneumonia detection using data Shapley values. Second, we show that low Shapley value indicates mislabeled examples, while high Shapley value indicates data points that are valuable for pneumonia detection. Finally, our approach can serve as a framework for using data Shapley to clean up large-scale medical imaging datasets.

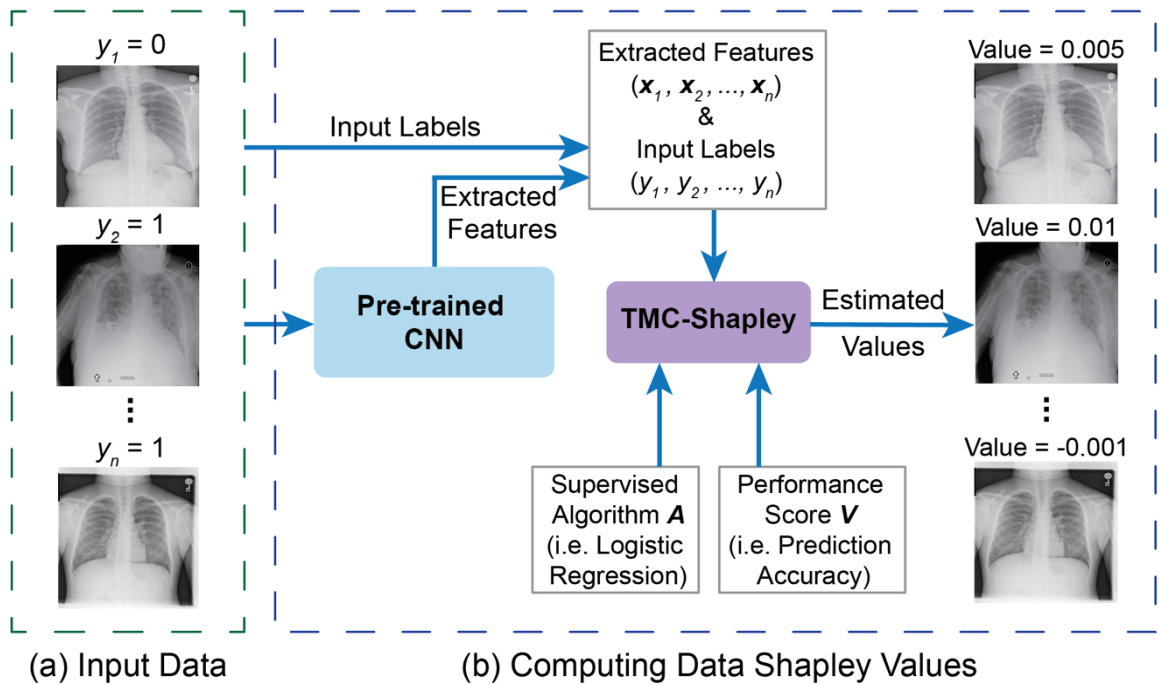

**Figure 1. Overview of our method.** (a) The input data were chest X-ray images and their corresponding labels (1 for pneumonia and 0 for no pneumonia) from ChestX-ray14 dataset[9]. (b) To compute data Shapley values for the training data, we first extracted feature vectors from a pre-trained convolutional neural network (CNN)[36]. Next, we applied TMC-Shapley[17]

to approximate the Shapley value of each training datum, where the supervised learning algorithm was logistic regression, and the predictor performance score was prediction accuracy for pneumonia.

# Results

*Overview*

In this study, we aim to characterize the effectiveness of data Shapley in identifying low quality and valuable data in ChestX-ray14, a large public chest X-ray dataset whose pathology labels were extracted from X-ray reports using text mining techniques. We sampled 2,000 chest X-rays as the training set to train the data Shapley algorithm and compute the Shapley values, 500 chest X-rays as the validation set to compute the predictor performance during training, and 610 chest X-rays as the held-out test sets to report the final results (see Table 1). We extracted features from a pre-trained convolutional neural network (CNN)[36] and computed the data Shapley value of each training datum with respect to the accuracy of a logistic regression algorithm for pneumonia detection. Furthermore, in collaboration with three radiologists, we evaluated the most valuable and least valuable chest X-rays, and provided qualitative interpretations for their Shapley values in the next sections.

**Table 1. Training, validation and held-out test sets used in our study**. All three sets were sampled from the train or validation set of ChestX-ray14 dataset[9]. The training set was used to train data Shapley algorithm and compute Shapley values, the validation set was used to compute the predictor performance during training, and the held-out test set was used to report the final results. Because the distribution of labels in the ChestX-ray14 dataset is highly imbalanced, we sampled a larger proportion of pneumonia cases in the training set and sampled balanced validation and held-out test sets in this study.

|  | Sample Size | # Pneumonia (%) |
| --- | --- | --- |
| **Training Set** | 2,000 | 200 (10.0%) |
| **Validation Set** | 500 | 249 (49.8%) |
| **Held-out Test Set** | 610 | 306 (50.2%) |

*Data Shapley identifies data points important for pneumonia detection*

Supplementary Figure S1 shows the histogram of data Shapley values for the training data. 41.5% of the training images had negative Shapley values. After calculating data Shapley values, we removed data points from the training set, starting from the most valuable datum to the least valuable, and trained a new logistic regression model each time when 1% training data were removed. Figure 2a-2c show the changes in prediction accuracy, precision and recall as data points were removed. Removing data points that have high Shapley values decreased the model performance. In contrast, removing data points that have high LOO values or randomly removing data points had little effect on the model performance. This suggests that data Shapley value is a highly accurate measure of a data point's importance since data points with high Shapley values were crucial to pneumonia detection. Interestingly, the 100 most valuable images were all labeled as pneumonia in ChestX-ray14, which might be because the training data were highly imbalanced.

Conversely, we removed data points from the least valuable to the most valuable as shown in Figure 2d-2f. Importantly, removing data points that have low Shapley values improved the model performance, suggesting that low Shapley value data points actually harm the model performance. Similarly, removing data points that have low LOO values or randomly removing data points did not affect the prediction accuracy or recall. We note that removing low Shapley value data points had a smaller effect on the precision score (Figure 2e), which might be because of the imbalanced training set. Besides the TMC-Shapley approximation method, we also explored a faster approximation, G-Shapley[17], and obtained comparable results (see Supplementary Figure S2).

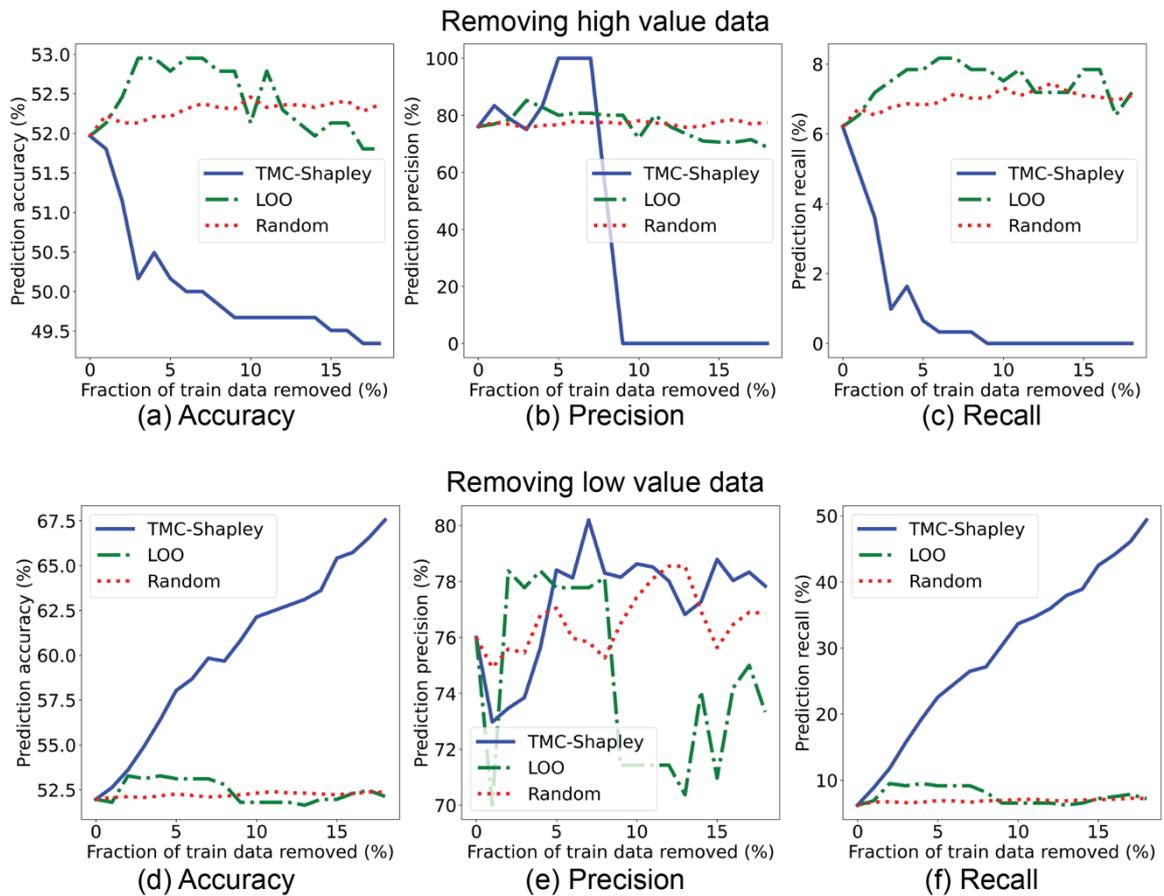

**Figure 2. (a) - (c) Effects of removing high value data points to pneumonia detection performance**. We removed the most valuable data points from the training set, as ranked by TMC-Shapley, leave-one-out (LOO) and uniform sampling (random) methods. We trained a new logistic regression model every time when 1% training data points were removed. The x-axis shows the percentage of training data removed, and the y-axis shows the model performance on the held-out test set in terms of (a) accuracy, (b) precision and (c) recall. Removing the most valuable data points identified by TMC-Shapley method decreased the model performance more than using LOO or randomly removing data. **(d) - (f) Effects of removing low value data points to pneumonia detection performance.** Conversely, we removed the least valuable data points from the training set. Removing the least valuable data points identified by TMC-Shapley method improved the model performance in terms of prediction (d) accuracy, (e) precision and (f) recall.

*Low Shapley value indicates mislabels in the dataset*

We asked three radiologists to re-label the 100 most valuable, 100 least valuable and 100 randomly sampled chest X-ray images in the training set. Supplementary Figure S3 shows the histograms of the Shapley values in these three sets of images. Next, we took the majority vote of the three radiologists' labels as the final label for each of these 300 images. Table 2 summarizes the radiologists' labels for the 300 images, and Supplementary Table S1 shows the inter-reader agreement. If no agreed label was achieved for an image, we excluded the image from further analyses (see Supplementary Table S1).

**Table 2. Three radiologists' relabeling results for the 100 least valuable, 100 most valuable and 100 randomly sampled chest X-ray images in the training set.** We used the majority vote to obtain the final label of each image. Disagreed images were excluded from further analyses. There were many more mislabeled examples in low value images (i.e. 65) than high value images (i.e. 22, pairwise p<0.0001) or random images (i.e. 20, pairwise p<0.0001), suggesting that low Shapley value effectively captures mislabels in the dataset.

|  | 100 Least Valuable Images | 100 Most Valuable Images | 100 Random Images | p value[a] |
|---|---|---|---|---|
| **# Originally Labeled as Pneumonia** | 13 | 100 | 5 | <0.0001 |
| **# Mislabels[b]** | 65 | 22 | 20 | <0.0001 |
| **# Mislabeled as Pneumonia** | 13 | 22 | 4 | 0.00078 |
| **# Mislabeled as No Pneumonia** | 52 | 0 | 16 | <0.0001 |

[a]p values computed using $\chi^2$ test.
[b]Note that since our training set has a higher percentage of pneumonia labels, the mislabel rates may not be representative for the entire ChestX-ray14 dataset.

There were two important observations from Table 2. First, there were many more mislabels in low value images (i.e. 65) compared to high value (i.e. 22, pairwise p<0.0001) or randomly sampled images (i.e. 20, pairwise p<0.0001). In particular, 13 low value images were mislabeled as pneumonia, and 52 were mislabeled as no pneumonia. This suggests that low Shapley value effectively captures mislabels in the dataset. Second, despite having high Shapley values, 22 images were mislabeled as pneumonia, which suggests that there might be other factors contributing to their high values (see further analyses in next sections). Note that no high value images were mislabeled as no pneumonia because all of the 100 most valuable images were associated with pneumonia in the dataset.

In addition, Supplementary Figure S4 visualizes the cumulative number of mislabels as data points were inspected in the descending, ascending and random orders of their Shapley values for the 100 most valuable, 100 least valuable and 100 randomly sampled images respectively. Low value images had a much steeper slope of accumulated mislabels compared to high value or randomly sampled images. In contrast, high value and randomly sampled images had similar number of mislabels, and their slopes of accumulated mislabels were similar.

*Heatmaps suggest contradictions between feature vectors and labels in low value images*

To better understand the factors contributing to low Shapley values, we visualized the most salient local regions in the 65 mislabeled low value chest X-ray images (see Methods). Figure 3a-3b show example heatmaps of low value chest X-rays that were mislabeled as pneumonia (Figure 3a) or no pneumonia (Figure 3b).

In Figure 3a, the heatmaps had low activations in relevant areas in the lung but high activations in irrelevant areas outside the lung, which suggests that the corresponding feature vectors favored no pneumonia over pneumonia for these images. Unsurprisingly, this pattern existed in the heatmaps of 12 (out of 13) low value images that were mislabeled as pneumonia.

In contrast, the heatmaps in Figure 3b had high activations in areas that indicate pneumonia, suggesting that their corresponding feature vectors favored pneumonia over no pneumonia. This pattern existed in the heatmaps of 45 (out of 52) low value images that were mislabeled as no pneumonia.

Therefore, the contradictions between feature vectors and labels likely resulted in these 65 mislabeled images being assigned low value.

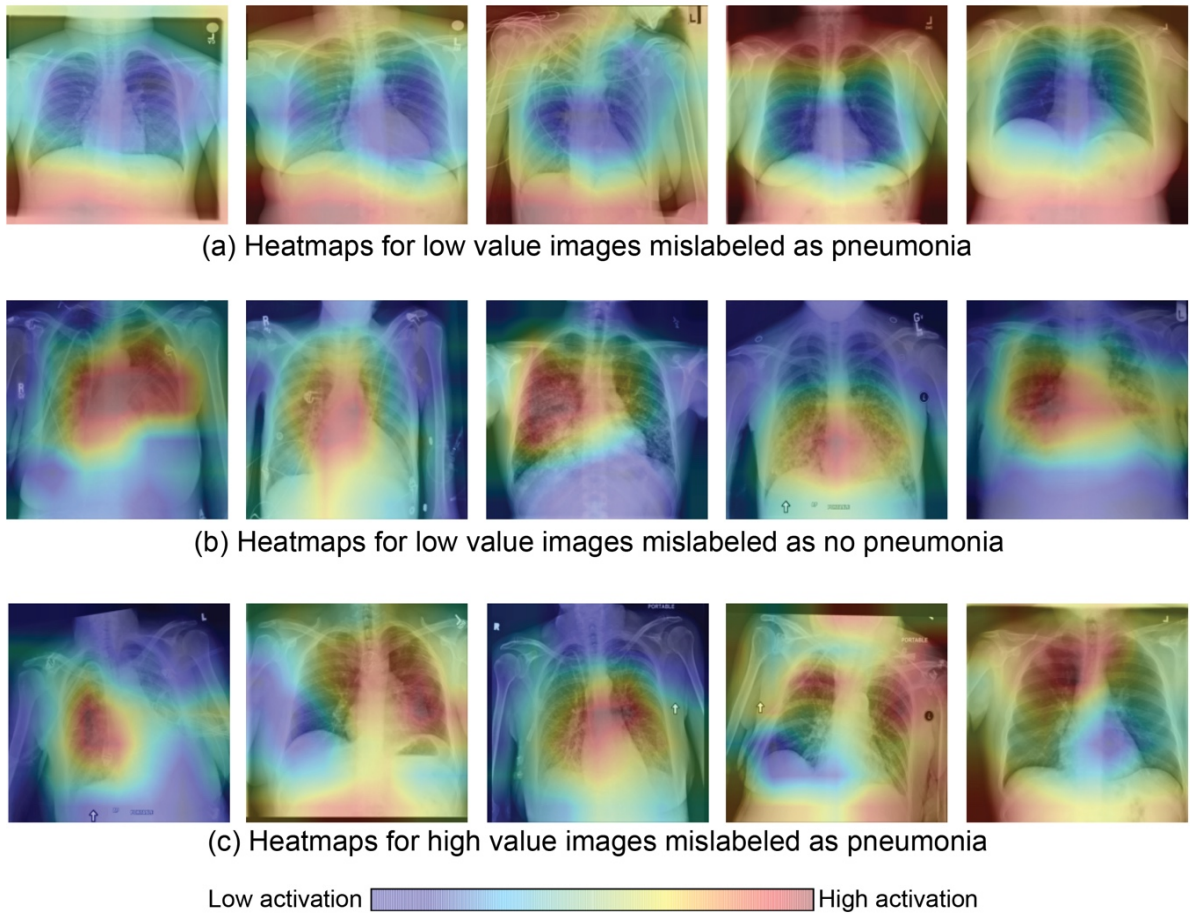

**Figure 3. Example heatmaps for (a) low value images that were mislabeled as pneumonia, (b) low value images that were mislabeled as no pneumonia, (c) high value images that were mislabeled as pneumonia.** (a) Heatmaps show low activations in relevant areas in the lung but high activations in irrelevant areas outside the lung, suggesting that the input feature vectors favored no pneumonia over pneumonia. (b) Heatmaps show high activations in lung areas indicating pneumonia, suggesting that the input feature vectors favored pneumonia over no pneumonia. (c) Heatmaps show high activations in abnormal lung areas. The first three images show abnormal opacity. The fourth image shows interstitial patterns, predominantly in the upper lung fields. The last image shows abnormal mass in the upper right lung field (i.e. upper left of the image).

*Abnormalities in chest X-rays are useful for pneumonia detection*

To understand why the 22 mislabeled chest X-ray images have high Shapley values (Table 2), we looked for other types of abnormalities in these images, and compared them to the 65 mislabeled low value images. The results are summarized in Table 3.

**Table 3. Number of mislabeled images that showed abnormalities among the 100 most valuable and 100 least valuable chest X-ray images**. All high value images that were mislabeled as pneumonia showed abnormalities to certain degrees. In contrast, only 2 out of 13 low value images that were mislabeled as pneumonia showed abnormalities. Moreover, 50 out of 52 low value images that were mislabeled as no pneumonia showed abnormal opacity.

|  | 22 High Value Images (Mislabeled as Pneumonia) | 13 Low Value Images (Mislabeled as Pneumonia) | 52 Low Value Images (Mislabeled as No Pneumonia) | p value[a] |
|---|---|---|---|---|
| # Images with Abnormalities | 22 | 2 | 50 | <0.0001 |

[a]p value computed using $\chi^2$ test.

Among the 22 mislabeled high value images, 10 images showed abnormal opacity (i.e. an important indicator for pneumonia) in the lung. Furthermore, the rest of the 12 images showed other kinds of abnormalities to certain degrees, including interstitial patterns and mass. Figure 3c shows example heatmaps of mislabeled high value images, where high activations correspond to abnormal areas. In contrast, only 2 out of the 13 low value images that were mislabeled as pneumonia had abnormalities, while the rest 11 images were completely normal. Moreover, among the 52 low value images that were mislabeled as no pneumonia, 50 images showed abnormal opacity in the lung. See Supplementary Figure S5 for examples of abnormalities in the mislabeled images.

Since there were relatively few true pneumonia cases in the training set, images mislabeled as pneumonia that showed abnormalities were still useful for the pneumonia detection algorithm. This was reflected in their Shapley values and could explain why the 22 mislabeled images had high Shapley values.

**Discussion**

Unlike prior studies that implicitly handle noisy labels or images in model development or training stages, our method directly identifies low value data points regardless of the reason for their poor quality. In addition to low value data, our method also informs us of data points that are valuable to pneumonia detection. Importantly, all of the 100 most valuable chest X-rays were labeled as pneumonia in ChestX-ray14. Since our training set is highly imbalanced (pneumonia versus no pneumonia ratio = 1:9), it is unsurprising that these rare positive data points are more important for pneumonia detection than the abundant negative data points.

Among the 100 chest X-ray images randomly sampled from our training set, 20 images (i.e. 20%) were mislabeled in ChestX-ray14 (see Table 2). Although this percentage might not represent the mislabeling rate in the entire ChestX-ray14 dataset, it suggests that a large portion of images in the dataset might not be reliably labeled. By visually inspecting images in ChestX-ray14, a recent study has shown that the pneumonia labels in the dataset only have 50% positive predictive value (PPV), lower than the 66% PPV reported in the original documentation of the dataset[12]. Our study further provides evidence towards the quality of the labels in ChestX-ray14. Since there are no gold-standard labels in ChestX-ray14, we suggest that users should be cautious about interpreting results from machine learning models evaluated on ChestX-ray14 data.

In our experiments, 22 images that were mislabeled as pneumonia were assigned high Shapley values (see Table 2). There are two possible reasons that resulted in this unexpected scenario. First, our analyses have shown that all of these images contain abnormalities to varying degrees. Hence, the feature vectors extracted from the pre-trained CNN might not have sufficient distinguishable representations between non-pneumonia and pneumonia abnormalities, and using representations from a better pre-trained model might mitigate this problem. Second, the logistic regression algorithm used for training data Shapley might have confused image features seen in non-pneumonia cases with those in pneumonia cases, and thus a more complex learning algorithm might be needed.

There are several limitations in our study. First, we did not use the full ChestX-ray14 dataset due to limited computational resources. As a future work, we can predict the Shapley values of the rest of the data points in ChestX-ray14 using the recently developed distributional Shapley framework[18], which would be more computationally efficient than direct computation of Shapley values. Second, since the ChestX-ray14 dataset has extremely imbalanced class labels, we sampled a larger percentage of pneumonia cases in our training, validation and held-out test sets (see Table 1). In order to accurately compute data Shapley values in imbalanced datasets, methods to explicitly account for class imbalance are needed. Third, because many of the abnormal chest X-rays show more than one thoracic diseases, it was hard for the model and the radiologists to tell for certain whether there is underlying pneumonia or not. Future works on predicting multi-class labels might better capture low quality data. Lastly, to show that our method can be used as a general approach for cleaning up massive medical datasets, validations on other medical data types are required.

Our approach has several important advantages. First, it allows us to have cleaner datasets and improves the performance of pneumonia detection. Second, it provides us insights into what types of chest X-rays are useful or harmful for pneumonia detection, which can inform us of the type of data to acquire to most efficiently improve the model performance. Third, our analyses suggest that not all mislabeled images are harmful --- some images that were originally mislabeled in ChestX-ray14 are still useful for pneumonia detection. Finally, our method can be easily extended for other medical data types such as time-series signals and electronic health records. For instance, one promising future direction is to use data Shapley values to prioritize what data to analyze for clinicians and thus help accelerate clinical workflows.

In conclusion, we used data Shapley to quantify the value of training data in a large-scale chest X-ray dataset in the context of pneumonia detection. We provided quantitative and qualitative analyses of the data Shapley values. We showed that low Shapley value indicates mislabels in the dataset, while high Shapley value indicates data points that are valuable for pneumonia detection. Our results are likely generalizable to other use cases, and our approach can serve as a framework for using data Shapley to denoise massive medical imaging datasets to improve the reliability of machine learning algorithms trained on such datasets.

# Methods

*Data*

In this study, we used ChestX-ray14[9], a large public chest X-ray dataset whose pathology labels were text-mined from X-ray reports. We chose ChestX-ray14 because it is known to have inaccurate labels[12]. As a proof of concept, we used 2,000 chest X-rays in ChestX-ray14 as the training set to train the data Shapley algorithm and compute the Shapley values, 500 chest X-rays as the validation set to compute the predictor performance score during training, and 610 chest X-rays as the held-out test set to report the final results. All of the 3,110 chest X-rays were sampled from the train or validation set of ChestX-ray14. Since only 1% of the chest X-rays in the dataset are associated with pneumonia, we sampled the training set with a larger proportion of pneumonia cases. In addition, for the ease of the interpretation of results (e.g. prediction accuracy, precision and recall), we sampled balanced validation and held-out test sets. Table 1 summarizes our data.

*Data valuation using Shapley value*

Shapley value has been widely used in the cooperative game theory literature as a fair measure for assigning contribution to each of the players in the game[37]. One can model the training of a machine learning model as a cooperative game where individual data points are the players. Therefore, Shapley value can be used for computing the contribution of each data point to the model's final performance. For a given set of training data points $D$ and a performance metric $V$ (e.g. test accuracy), The "Data Shapley" value $\phi_i$ of a data point $x_i \in D$ is defined as[17]:

$$\phi_i = \sum_{S \subseteq D \setminus \{x_i\}} \frac{V(S \cup \{x_i\}) - V(S)}{\binom{|D|-1}{|S|}} \quad (1)$$

where $V(S)$ is the performance of the model trained on subset $S$ of the data. In our setting, $V(S)$ is the pneumonia prediction accuracy on the validation set. Intuitively, the Shapley value of a data point is a weighted average of its marginal contribution to subsets of the rest of the dataset. As a result, it can be used as a measure of data quality: a data point with a high Shapley value is one that improves the model's performance if we add it to most subsets of the data, while a data point with a negative value on average hurts the performance of the model. Exact computation of Equation (1) requires an exponential number of computations in the size of the dataset, which is infeasible in most realistic settings. Previous work[17] has developed an efficient approximation method called "TMC-Shapley" which has shown promising performance. TMC-Shapley is efficient based on two approximations: (a) by rearranging the above formula, it can be shown that the data Shapley values are expectations of bounded random variables, which can be approximated by Monte-Carlo sampling; (b) the marginal contribution of a point $x_i \in D$ is close to zero when it is added to a large subset, i.e. $V(S \cup \{x_i\}) - V(S) \approx 0$ for a large enough $|S|$.

*Quantifying the value of chest X-rays with data Shapley*

An overview of our method to compute data Shapley values for the chest X-rays is shown in Figure 1. First, we extracted features from the last average-pooling layer (i.e. before the last fully connected layer) of a pre-trained CNN[36], which resulted in a 1,024-dimensional feature vector for each chest X-ray. Next, we applied the TMC-Shapley[17] algorithm to calculate the value for each chest X-ray for pneumonia detection. Specifically, we denote the training data as $D = \{x_i, y_i\}_{i=1}^{n}$ where $n$ was the size of the training set, $x_i \in \mathbb{R}^{1,024}$ was the feature vector, and $y_i \in \{0, 1\}$ was the pneumonia label (0 for no pneumonia and 1 for pneumonia). We used logistic regression as the supervised learning algorithm and prediction accuracy as the performance metric $V$. As a comparison to data Shapley, we also computed the leave-one-out (LOO) value[35] for each training datum. After calculating data Shapley values, we removed the most (least) valuable data points from the training set, as ranked by Shapley values, LOO values and uniform sampling (i.e. random). We trained a new logistic regression model every time when data points were removed. Moreover, we investigated the 100 least valuable and the 100 most valuable chest X-rays in the training set based on their Shapley values. In addition, we randomly sampled another 100 chest X-rays from the training set as a comparison to the 100 most valuable and 100 least valuable images.

*Investigating relations between data Shapley values and input labels*

First, we investigated the relations between the Shapley values and the labels. We asked three radiologists (R.Y., S.R. and D.L.R.) to re-label 300 chest X-rays (i.e. 100 least valuable, 100 most valuable and 100 randomly sampled from the training set). The radiologists were blinded to the original labels and the Shapley values while labeling the chest X-rays. For each chest X-ray, the radiologists had the choice of labeling it as pneumonia, no pneumonia or unsure.

In order to further evaluate mislabeled images and their Shapley values, we asked the three radiologists to re-evaluate the 100 most valuable and the 100 least valuable chest X-ray images, and determined whether these images have other types of abnormalities (e.g. opacity) or are completely normal.

*Visualizing heatmaps for chest X-rays*

To better understand the mislabeled chest X-ray images, we visualized heatmaps that show the local regions in the images leading to the prediction of pneumonia[38]. Specifically, we fed a chest X-ray image into the same pre-trained CNN[36] and took the output of the final convolutional layer as the feature maps. Let $f_k \in \mathbb{R}^{7\times 7}$ be the $k$-th feature map and $w_k$ be the final classification layer weight for the $k$-th feature map. Taking the weighted sum of the feature maps and the corresponding weights, we obtained a heatmap $M \in \mathbb{R}^{7\times 7}$ that shows the most salient features for the prediction of pneumonia. Mathematically, the heatmap $M$ was computed as follows:

$$M = \sum_{k=1}^{1,024} w_k f_k \qquad (2)$$

Lastly, we overlaid the heatmap with the original image, which allowed us to visualize the most salient local regions in the image leading to the prediction of pneumonia.

*Statistical tests*

$\chi^2$ tests were used to compare the proportions of mislabels in the three sets of chest X-rays (i.e. 100 most valuable, 100 least valuable and 100 randomly sampled), as well as the proportions of abnormalities in mislabeled images among the 100 most valuable and 100 least valuable chest X-rays.

## Data Availability

The chest X-rays used for this study are publicly available at https://nihcc.app.box.com/v/ChestXray-NIHCC. The code for pre-training the convolutional neural network is available at https://github.com/jrzech/reproduce-chexnet.

The list of images in our training, validation and held-out test sets can be obtained from the authors upon request.

# References

1. Ouyang, D. *et al.* Video-based AI for beat-to-beat assessment of cardiac function. *Nature* **580**, 252–256 (2020).

2. Ronneberger, O., Fischer, P. & Brox, T. U-Net: Convolutional Networks for Biomedical Image Segmentation. in *Medical Image Computing and Computer-Assisted Intervention -- MICCAI 2015* 234–241 (Springer International Publishing, 2015).

3. Esteva, A. *et al.* Dermatologist-level classification of skin cancer with deep neural networks. *Nature* **542**, 115–118 (2017).

4. Titano, J. J. *et al.* Automated deep-neural-network surveillance of cranial images for acute neurologic events. *Nat. Med.* **24**, 1337–1341 (2018).

5. Lee, H. *et al.* An explainable deep-learning algorithm for the detection of acute intracranial haemorrhage from small datasets. *Nat. Biomed. Eng.* **3**, 173 (2019).

6. Rajpurkar, P. *et al.* Deep learning for chest radiograph diagnosis: A retrospective comparison of the CheXNeXt algorithm to practicing radiologists. *PLoS Med.* **15**, e1002686 (2018).

7. Dunnmon, J. A. *et al.* Assessment of convolutional neural networks for automated classification of chest radiographs. *Radiology* **290**, 537–544 (2019).

8. Esteva, A. *et al.* A guide to deep learning in healthcare. *Nat. Med.* **25**, 24–29 (2019).

9. Wang, X. *et al.* ChestX-ray8: Hospital-scale chest X-ray database and benchmarks on weakly-supervised classification and localization of common thorax diseases. *Proc. - 30th IEEE Conf. Comput. Vis. Pattern Recognition, CVPR 2017* **2017-Janua**, 3462–3471 (2017).

10. Yan, K., Wang, X., Lu, L. & Summers, R. M. DeepLesion: automated mining of large-scale lesion annotations and universal lesion detection with deep learning. *J. Med. Imaging* **5**, 1 (2018).

11. Johnson, A. E. W. *et al.* MIMIC-CXR, a de-identified publicly available database of chest radiographs with free-text reports. *Sci. Data* **6**, 317 (2019).

12. Oakden-Rayner, L. Exploring Large-scale Public Medical Image Datasets. *Acad. Radiol.* **27**, 106–112 (2020).

13. Gurari, D. *et al.* How to collect segmentations for biomedical images? A benchmark evaluating the performance of experts, crowdsourced non-experts, and algorithms. in *2015 IEEE winter conference on applications of computer vision* 1169–1176 (IEEE,

2015).

14. Willemink, M. J. *et al.* Preparing medical imaging data for machine learning. *Radiology* **295**, 4–15 (2020).

15. van Ooijen, P. M. A. Quality and Curation of Medical Images and Data BT - Artificial Intelligence in Medical Imaging: Opportunities, Applications and Risks. in 247–255 (Springer International Publishing, 2019). doi:10.1007/978-3-319-94878-2_17

16. Zech, J. R. *et al.* Variable generalization performance of a deep learning model to detect pneumonia in chest radiographs: a cross-sectional study. *PLoS Med.* **15**, e1002683 (2018).

17. Ghorbani, A. & Zou, J. Data Shapley: Equitable Valuation of Data for Machine Learning. in *International Conference on Machine Learning* 2242–2251 (2019).

18. Ghorbani, A., Kim, M. P. & Zou, J. A Distributional Framework for Data Valuation. *Int. Conf. Mach. Learn.* (2020).

19. Karimi, D., Dou, H., Warfield, S. K. & Gholipour, A. Deep learning with noisy labels: exploring techniques and remedies in medical image analysis. *arXiv Prepr. arXiv1912.02911* (2019).

20. Dgani, Y., Greenspan, H. & Goldberger, J. Training a neural network based on unreliable human annotation of medical images. in *2018 IEEE 15th International Symposium on Biomedical Imaging (ISBI 2018)* 39–42 (IEEE, 2018).

21. Le, H. *et al.* Pancreatic cancer detection in whole slide images using noisy label annotations. in *International Conference on Medical Image Computing and Computer-Assisted Intervention* 541–549 (Springer, 2019).

22. Xue, C., Dou, Q., Shi, X., Chen, H. & Heng, P.-A. Robust learning at noisy labeled medical images: applied to skin lesion classification. in *2019 IEEE 16th International Symposium on Biomedical Imaging (ISBI 2019)* 1280–1283 (IEEE, 2019).

23. Yu, L., Wang, S., Li, X., Fu, C.-W. & Heng, P.-A. Uncertainty-aware self-ensembling model for semi-supervised 3D left atrium segmentation. in *International Conference on Medical Image Computing and Computer-Assisted Intervention* 605–613 (Springer, 2019).

24. Fries, J. A. *et al.* Weakly supervised classification of aortic valve malformations using unlabeled cardiac MRI sequences. *Nat. Commun.* **10**, 1–10 (2019).

25. Dunnmon, J. A. *et al.* Cross-modal data programming enables rapid medical machine learning. *Patterns* 100019 (2020).

26. Calli, E., Sogancioglu, E., Scholten, E. T., Murphy, K. & van Ginneken, B. Handling

label noise through model confidence and uncertainty: application to chest radiograph classification. in *Medical Imaging 2019: Computer-Aided Diagnosis* **10950**, 1095016 (International Society for Optics and Photonics, 2019).

27. Aviles-Rivero, A. I. *et al.* GraphXNET- Chest X-Ray Classification Under Extreme Minimal Supervision. in *International Conference on Medical Image Computing and Computer-Assisted Intervention* 504–512 (Springer, 2019).

28. Bhonsle, D., Chandra, V. & Sinha, G. R. Medical image denoising using bilateral filter. *Int. J. Image, Graph. Signal Process.* **4**, 36 (2012).

29. Kaur, P., Singh, G. & Kaur, P. A Review of Denoising Medical Images Using Machine Learning Approaches. *Curr. Med. Imaging Rev.* **14**, 675–685 (2018).

30. Rabbani, H., Nezafat, R. & Gazor, S. Wavelet-domain medical image denoising using bivariate laplacian mixture model. *IEEE Trans. Biomed. Eng.* **56**, 2826–2837 (2009).

31. Wang, Y. & Zhou, H. Total variation wavelet-based medical image denoising. *Int. J. Biomed. Imaging* **2006**, (2006).

32. Gondara, L. Medical image denoising using convolutional denoising autoencoders. in *2016 IEEE 16th International Conference on Data Mining Workshops (ICDMW)* 241–246 (IEEE, 2016).

33. Jifara, W., Jiang, F., Rho, S., Cheng, M. & Liu, S. Medical image denoising using convolutional neural network: a residual learning approach. *J. Supercomput.* **75**, 704–718 (2019).

34. Gong, K., Guan, J., Liu, C.-C. & Qi, J. PET image denoising using a deep neural network through fine tuning. *IEEE Trans. Radiat. Plasma Med. Sci.* **3**, 153–161 (2018).

35. Cook, R. D. Detection of influential observation in linear regression. *Technometrics* **19**, 15–18 (1977).

36. Rajpurkar, P. *et al.* Chexnet: Radiologist-level pneumonia detection on chest x-rays with deep learning. *arXiv Prepr. arXiv1711.05225* (2017).

37. Shapley, L. S. A value for n-person games. *Contrib. to Theory Games* **2**, 307–317 (1953).

38. Zhou, B., Khosla, A., Lapedriza, A., Oliva, A. & Torralba, A. Learning deep features for discriminative localization. in *Proceedings of the IEEE conference on computer vision and pattern recognition* 2921–2929 (2016).

## Acknowledgements

This work was supported by a grant from the Wu Tsai Neurosciences Institute. D.L.R. is supported in part by grants from the National Cancer Institute, 1U01CA190214, 1U01CA187947, U01CA242879, and U24CA226110. J.Z. is supported by NSF CCF 1763191, NSF CAREER 1942926, NIH P30AG059307, NIH U01MH098953 and grants from the Silicon Valley Foundation and the Chan-Zuckerberg Initiative.

**Author Contributions**

S.T., A.G., J.Z. and D.L.R. conceived and designed the study, S.T., R.Y., S.R. and D.L.R. analyzed the data, S.T. drafted the manuscript and all authors critically revised the manuscript for important intellectual content.

## Additional Information

The authors declare no competing interests.

# Data Valuation for Medical Imaging Using Shapley Value: Application on A Large-scale Chest X-ray Dataset

# Supplementary Information

**Supplementary Table S1.** Inter-reader agreement for (a) 100 least valuable images, (b) 100 most valuable images and (c) 100 randomly sampled images.

(a) 100 Least valuable images

|  | Pneumonia | No Pneumonia | Unsure |
|---|---|---|---|
| **# Agreed by Three Radiologists** | 10 | 18 | 0 |
| **# Agreed by Two Radiologists** | 42 | 14 | 6 |
| **# Disagreed by Three Radiologists** | 10 | | |

(b) 100 Most valuable images

|  | Pneumonia | No Pneumonia | Unsure |
|---|---|---|---|
| **# Agreed by Three Radiologists** | 24 | 7 | 0 |
| **# Agreed by Two Radiologists** | 41 | 15 | 7 |
| **# Disagreed by Three Radiologists** | 6 | | |

(c) 100 Randomly sampled images

|  | Pneumonia | No Pneumonia | Unsure |
|---|---|---|---|
| **# Agreed by Three Radiologists** | 3 | 64 | 0 |
| **# Agreed by Two Radiologists** | 13 | 15 | 1 |
| **# Disagreed by Three Radiologists** | 4 | | |

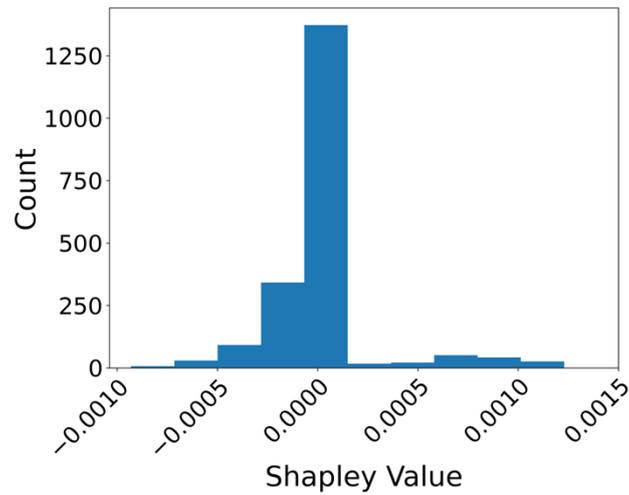

**Supplementary Figure S1. Histogram of Shapley values for training data.** Most data points have Shapley values around 0. Data points with negative Shapley values (41.5%) on average harm the performance of the pneumonia detection algorithm, whereas data points with high Shapley values are valuable for pneumonia detection.

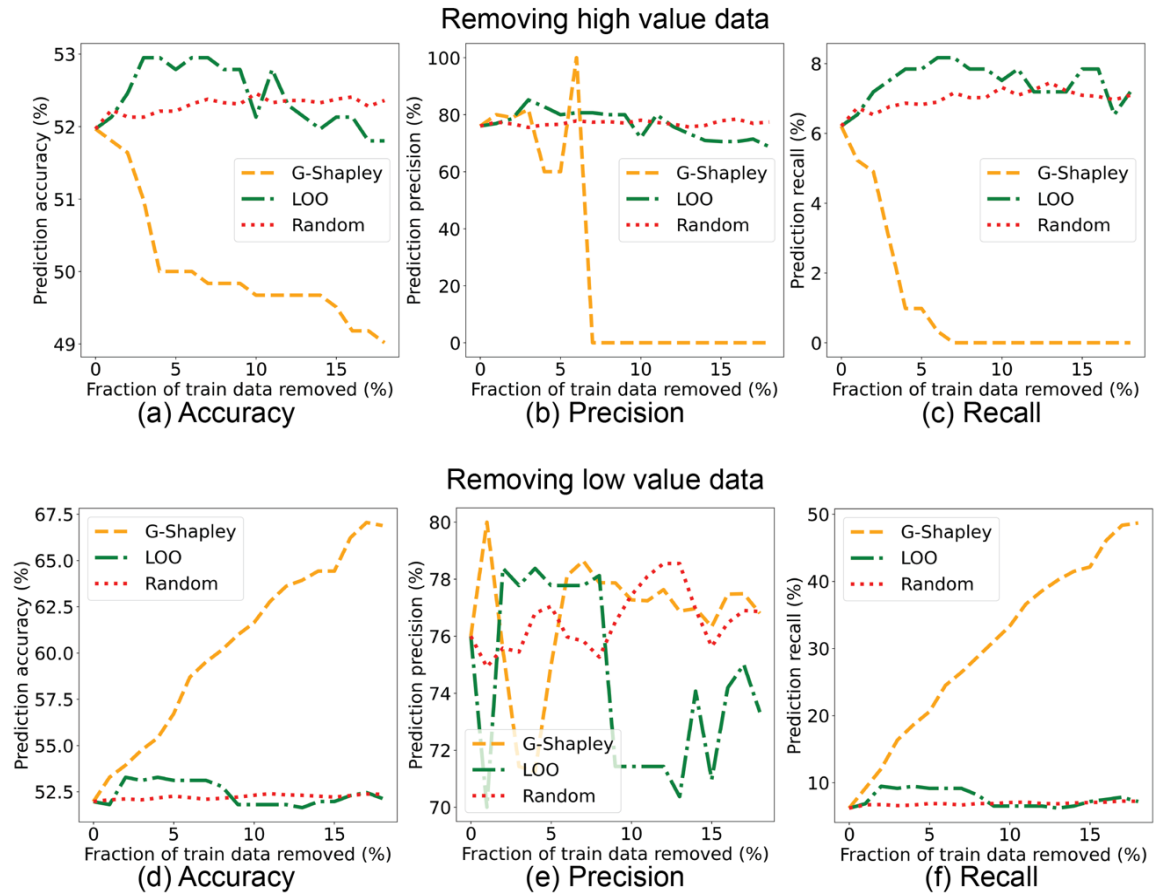

**Supplementary Figure S2.** In addition to TMC-Shapley approximation, we experimented with G-Shapley[1] to approximate data Shapley values and obtained similar results as that with TMC-Shapley (Figure 2). **(a) - (c) Effects of removing high value data points to pneumonia detection performance**. We removed the most valuable data points from the training set, as ranked by G-Shapley, leave-one-out (LOO) and uniform sampling (random) methods. We trained a new logistic regression model every time when 1% training data points were removed. The x-axis shows the percentage of training data removed, and the y-axis shows the model performance on the held-out set in terms of (a) accuracy, (b) precision and (c) recall. Removing the most valuable data points identified using G-Shapley method decreased the model performance more than using LOO or randomly removing data. **(d) - (f) Effects of removing low value data points to pneumonia detection performance.** Conversely, we removed the least valuable data points from the training set. Removing the least valuable data points identified by G-Shapley method greatly improved the model performance in terms of prediction (d) accuracy and (f) recall.

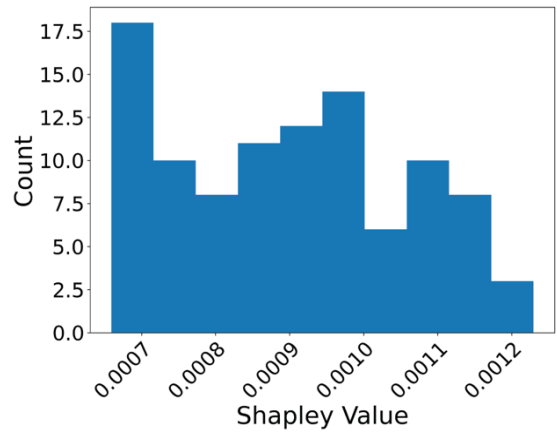

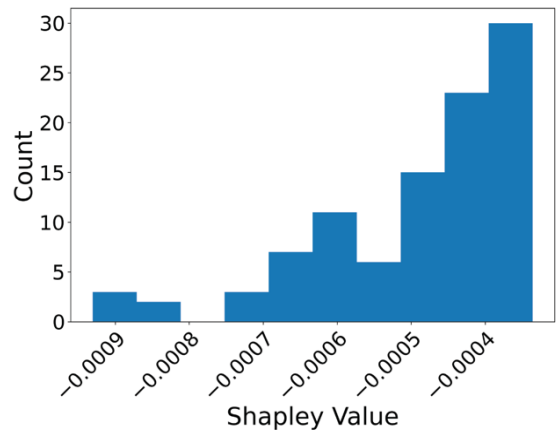

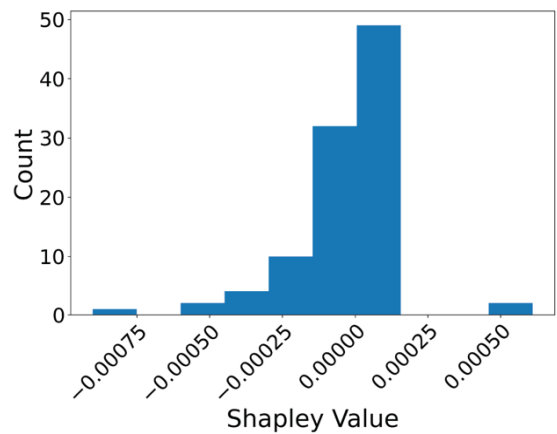

**Supplementary Figure S3. Histogram of Shapley values for** (a) 100 most valuable images, (b) 100 least valuable images and (c) 100 randomly sampled images in the training set.

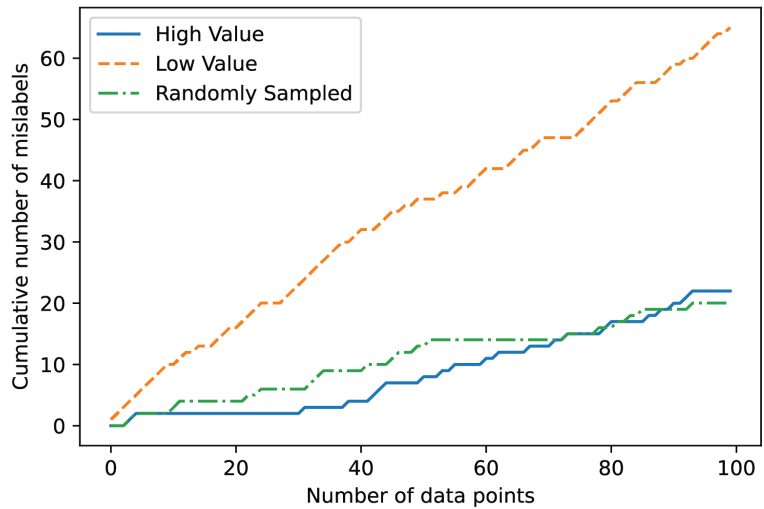

**Supplementary Figure S4. Cumulative number of mislabels in the 100 most valuable, 100 least valuable and 100 randomly sampled images in the training set.** The x-axis shows the number of data points visited. The y-axis shows the cumulative number of mislabels. For the 100 most valuable images (blue curve), the data points are sorted from the highest to the lowest Shapley value. For the 100 least valuable images (orange curve), the data points are sorted from the lowest to the highest Shapley value. For the 100 randomly sampled images (green curve), the data points are sorted randomly. Low value images had much more mislabels and a much steeper slope of accumulated mislabels compared to high value or randomly sampled images.

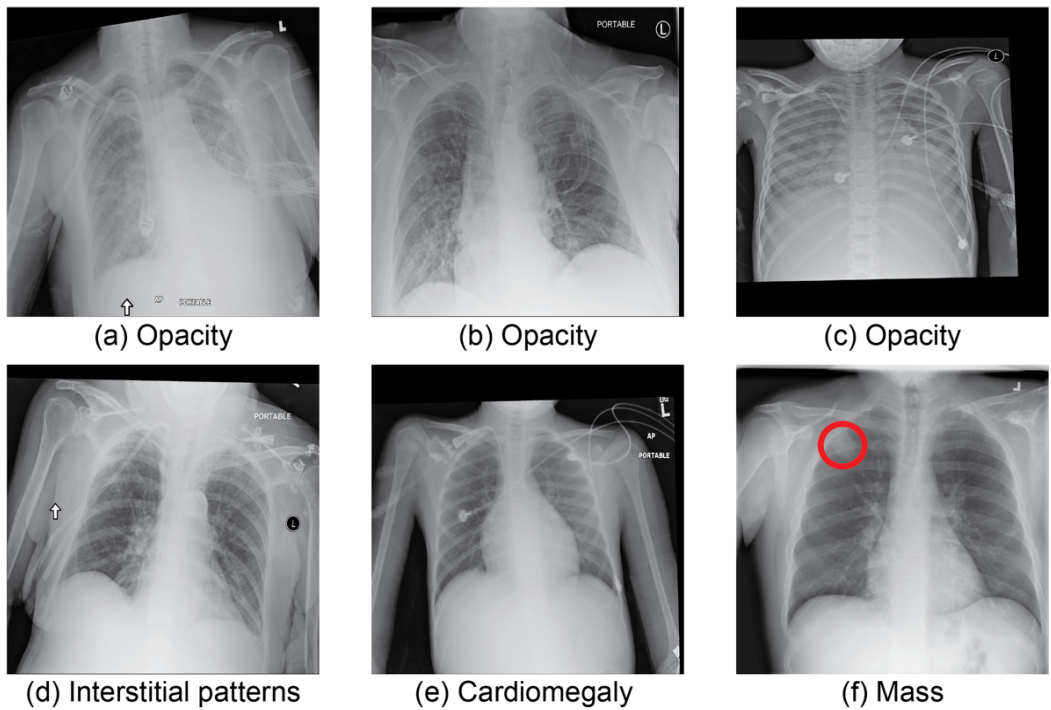

**Supplementary Figure S5. Example abnormalities observed in the mislabeled images.**
(a)-(c) Bilateral opacity in both lung fields. (d) Abnormal interstitial patterns that are prominent in the upper lung areas. (e) Cardiomegaly (i.e. abnormally enlarged heart). (f) Abnormal mass in the upper right lung (circled in red).

# References

1. Ghorbani, A. & Zou, J. Data Shapley: Equitable Valuation of Data for Machine Learning. in *International Conference on Machine Learning* 2242–2251 (2019).